\documentclass[10pt,twocolumn,letterpaper]{article}
\usepackage[T1]{fontenc}
\usepackage[utf8]{inputenc}
\usepackage[margin=1in,footskip=0.35in]{geometry}
\usepackage{times}
\usepackage{microtype}
\usepackage{graphicx,amsmath,amssymb,bm}
\usepackage{booktabs,multirow,array}
\usepackage[dvipsnames,table]{xcolor}
\usepackage{float}
\usepackage[numbers,sort&compress]{natbib}
\usepackage[format=plain,labelformat=simple,labelsep=period,font=small]{caption}
\usepackage[font=footnotesize,skip=3pt,subrefformat=parens]{subcaption}
\usepackage{enumitem}
\usepackage{titlesec}
\titleformat{\section}{\normalfont\large\bfseries}{\thesection}{0.6em}{}
\titleformat{\subsection}{\normalfont\normalsize\bfseries}{\thesubsection}{0.6em}{}
\titlespacing*{\section}{0pt}{2.0ex plus 0.6ex minus 0.2ex}{1.0ex plus 0.2ex}
\titlespacing*{\subsection}{0pt}{1.8ex plus 0.5ex minus 0.2ex}{0.8ex plus 0.2ex}
\setlist[itemize]{noitemsep,leftmargin=*,topsep=3pt}
\usepackage[hyphens]{url}
\usepackage[breaklinks,colorlinks,linkcolor=MidnightBlue,citecolor=MidnightBlue,urlcolor=MidnightBlue]{hyperref}
\usepackage[capitalize]{cleveref}
\crefname{section}{Sec.}{Secs.}
\Crefname{section}{Section}{Sections}
\crefname{table}{Tab.}{Tabs.}
\Crefname{table}{Table}{Tables}
\hypersetup{
  pdftitle={GeoTex: Geometry-as-Texture Encoding and History-Grounded 4D Occupancy Forecasting},
  pdfauthor={Shijun Guo, Yaming Yang, Haoran Xu, Wei Zhao, Ziyu Guan, Cai Xu, Peixi Peng},
  pdfsubject={4D occupancy forecasting},
  pdfkeywords={occupancy forecasting, geometric texture, history-grounded prediction}
}
\newcommand{\paneltitle}[1]{{\normalfont\rmfamily\footnotesize #1}}

\title{\LARGE\bfseries GeoTex: Geometry-as-Texture Encoding and History-Grounded\\4D Occupancy Forecasting}
\author{%
\normalsize Shijun Guo\textsuperscript{1}\quad
Yaming Yang\textsuperscript{1}\quad
Haoran Xu\textsuperscript{2}\quad
Wei Zhao\textsuperscript{1}\thanks{Corresponding author.}\\[0.4em]
\normalsize Ziyu Guan\textsuperscript{1}\quad
Cai Xu\textsuperscript{1}\quad
Peixi Peng\textsuperscript{2}\\[0.5em]
\normalsize \textsuperscript{1}Xidian University\qquad
\textsuperscript{2}Peking University\\[0.3em]
\small \texttt{ywzhao@mail.xidian.edu.cn}
}

\date{}
\begin{document}
\maketitle
\begin{abstract}
4D occupancy forecasting models 3D scene evolution to support future prediction and scenario generation in autonomous-driving world models. Semantic occupancy abstracts image color and texture into discrete labels, leaving 3D voxel arrangements as the primary cues for distinguishing local shapes within each class. Existing methods borrow image-encoding frameworks to obtain compact latent representations through height-channel encoding or plane factorization, but do not sufficiently model within-column height patterns and cross-column relations jointly before compression, weakening these geometric cues. Forecasting also underutilizes complementary history and struggles to retain reliable observation evidence. We therefore propose GeoTex, an occupancy world model combining geometry-as-texture encoding with history-grounded forecasting. GeoTex Representation jointly encodes within-column height patterns and cross-column relations into compact continuous latent representations and restores 3D structure through content-dependent height unfolding and geometric interaction. GeoTex Prediction retains initial evidence throughout rollout in fixed cross-observation relational memory and incorporates complementary recent history through target-referenced residual updates. After prediction training, decoder-only adaptation improves semantic decoding of predicted states. GeoTex improves average forecasting mIoU and IoU by \textbf{9.79\%} and \textbf{11.91\%} relative to GenieDrive while reducing parameters from 3.47M to 2.33M, with gains increasing at longer horizons.
\end{abstract}

\begin{figure}[!t]
    \centering
    \begin{minipage}{0.98\linewidth}
        \centering
        \offinterlineskip
        \includegraphics[width=\linewidth,trim=0 660bp 0 0,clip]{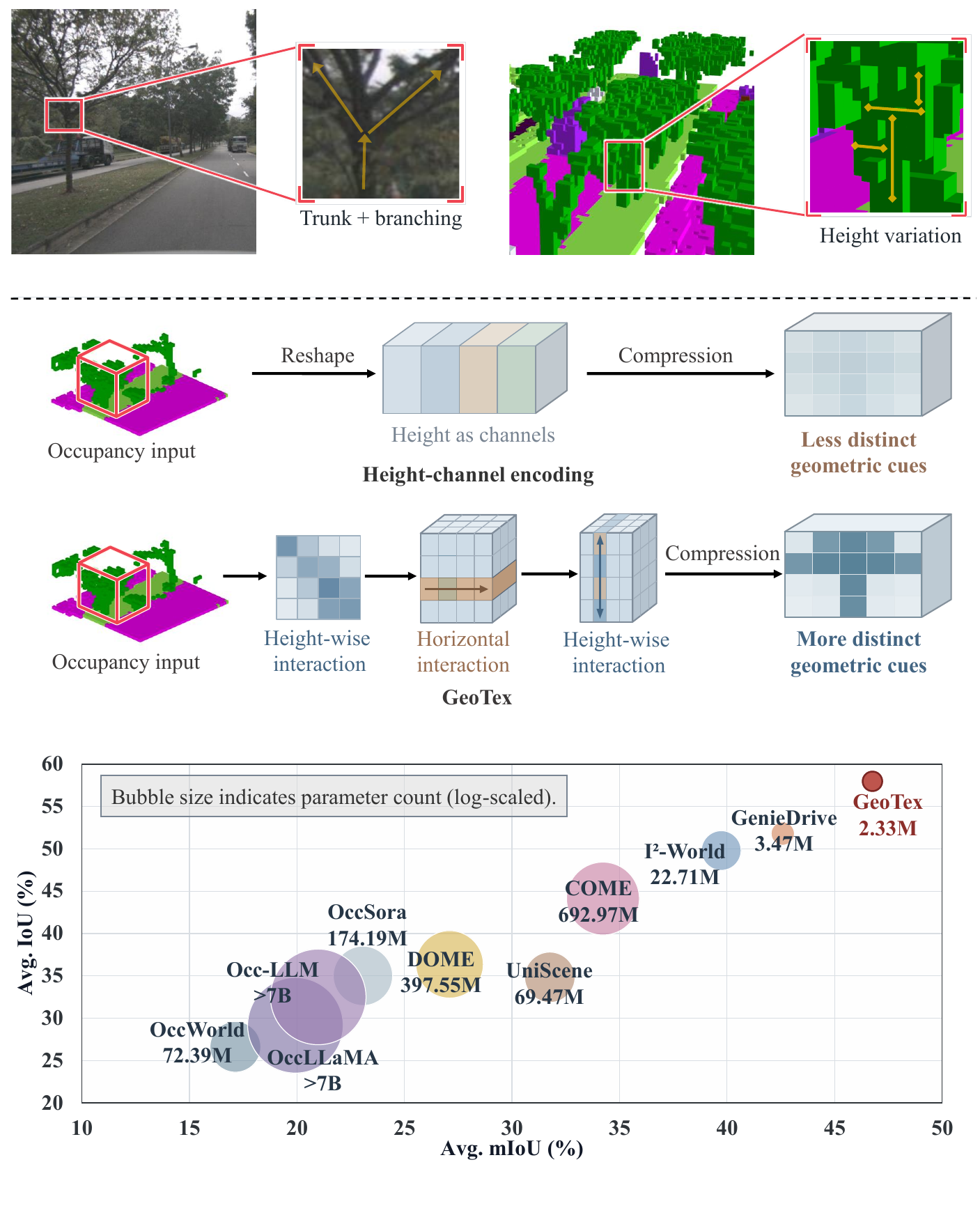}\par
        \vspace{3pt}
        \makebox[0.5\linewidth]{\paneltitle{(a) Image texture}}\makebox[0.5\linewidth]{\paneltitle{(b) Occupancy labels}}\par
        \vspace{2pt}
        \includegraphics[width=\linewidth,trim=0 346bp 0 202bp,clip]{fig/introduction/fig1_abcd_artwork.pdf}\par
        \vspace{3pt}
        \paneltitle{(c) Geometry-as-texture encoding}\par
        \vspace{2pt}
        \includegraphics[width=\linewidth,trim=0 35bp 0 516bp,clip]{fig/introduction/fig1_abcd_artwork.pdf}\par
        \vspace{3pt}
        \paneltitle{(d) Forecasting accuracy vs. model size}\par
    \end{minipage}
    \caption{\textbf{Geometry as texture and forecasting efficiency.} (a--b) Semantic occupancy expresses local shape through geometric arrangements rather than appearance texture. (c) GeoTex preserves height patterns and cross-column relations during compression. (d) GeoTex establishes a state-of-the-art accuracy--efficiency trade-off, attaining the highest average mIoU and IoU with only 2.33M parameters.}
    \label{fig:geometry_as_texture}
\end{figure}

\section{Introduction}
\label{sec:introduction}

Occupancy world models (OWMs)~\cite{occworld,i2world,geniedrive} represent driving environments with 3D semantic occupancy and model future scene evolution through 4D occupancy forecasting, supporting autonomous-driving planning~\cite{driveworld,driveoccworld,gwm} and simulation~\cite{wovogen,drivingsphere,xscene}. Occupancy representations unify geometry and semantics in semantic voxel grids~\cite{tpvformer,surroundocc}, capturing the 3D shapes and boundaries of irregular objects without relying on bounding boxes. To efficiently model scene evolution, these models encode historical occupancy into compact latent states and predict their changes. Forecasting quality therefore depends both on the scene structure preserved in latent representations and on how history is used during prediction.

To construct these states, existing OWMs primarily adopt VAE-series architectures from image generation~\cite{kingma2014vae,ldm,vqgan,rqvae,var}, but their encoding designs do not fully account for how occupancy expresses local shape. In \cref{fig:geometry_as_texture}(a--b), image color and texture provide cues for distinguishing tree structures~\cite{gatys2015texture,guidedtexture,selfrecttexture}, whereas semantic occupancy abstracts these variations into a shared vegetation label, making local shape discrimination depend on 3D voxel arrangements. Continuation, gaps, and termination along vertical columns define height patterns, while variations and connections across columns describe surfaces, boundaries, and spatial extent. Together, these relations form the \emph{geometric texture} of occupancy scenes. Height-channel encoding~\cite{occworld,i2world} and tri-plane factorization~\cite{eg3d,geniedrive} produce compact representations but insufficiently model these relations jointly before compression, weakening the geometric cues that distinguish local shape (\cref{fig:geometry_as_texture}(c)). This mismatch between encoding design and how occupancy expresses structure motivates encoding centered on occupancy geometry.

Forecasting also requires effective use of historical evidence. Existing methods learn scene evolution through autoregressive prediction~\cite{occworld,i2world,geniedrive} or diffusion generation~\cite{occsora,dome,come}, and some further combine geometric propagation and content refinement~\cite{occsim,efficientocf,gast} to exploit spatial correspondence for forecasting. However, historical feature aggregation does not explicitly distinguish what each frame adds relative to the current target; even with spatial correspondence, observations still differ in content due to visibility and sampling. Meanwhile, during autoregressive rollout with sliding history, real observations gradually leave the window, limiting continued access to initial observation evidence. Moreover, a decoder trained only on states encoded from ground-truth occupancy faces a distribution shift when decoding autoregressively generated states, affecting semantic decoding quality.

To address these issues, we propose \textbf{GeoTex}, an occupancy world model that combines geometry-as-texture encoding with history-grounded forecasting. At the representation stage, \textbf{GeoTex Representation} alternates height-wise and horizontal interactions after local 3D encoding, linking within-column shape to surrounding structures before height-ordered aggregation into compact continuous latent states. During reconstruction, the decoder unfolds content-dependent height features and restores occupancy through geometric interactions and 3D upsampling.

In this continuous latent space, \textbf{GeoTex Prediction} retains initial observation evidence throughout rollout in fixed relational memory and tracks recent states through sliding history. At each prediction step, ego-motion locates historical regions corresponding to target positions, and current target features guide content retrieval. To exploit complementary information from recent history, we introduce a \emph{target-referenced historical residual update} that subtracts current target features from retrieved historical features, weights the residuals across frames, and writes them into the target state. After prediction training, we freeze the other modules and adapt the decoder using encoded and predicted states to improve semantic decoding.

Our contributions are summarized as follows:
\begin{itemize}
    \item We introduce GeoTex Representation, which jointly organizes height patterns and cross-column relations during compression and reconstruction to preserve geometric texture in occupancy scenes.
    \item We propose GeoTex Prediction, retaining initial scene evidence in relational memory and extracting complementary updates through target-referenced historical residuals for history-grounded forecasting.
    \item With only \textbf{2.33M} parameters, GeoTex improves average forecasting mIoU and IoU by \textbf{9.79\%} and \textbf{11.91\%} relative to leading methods, better preserves scene geometry, and widens its advantage at longer horizons.
\end{itemize}

\section{Related Work}
\label{sec:related_work}

\begin{figure*}[!t]
    \centering
    \includegraphics[width=\textwidth]{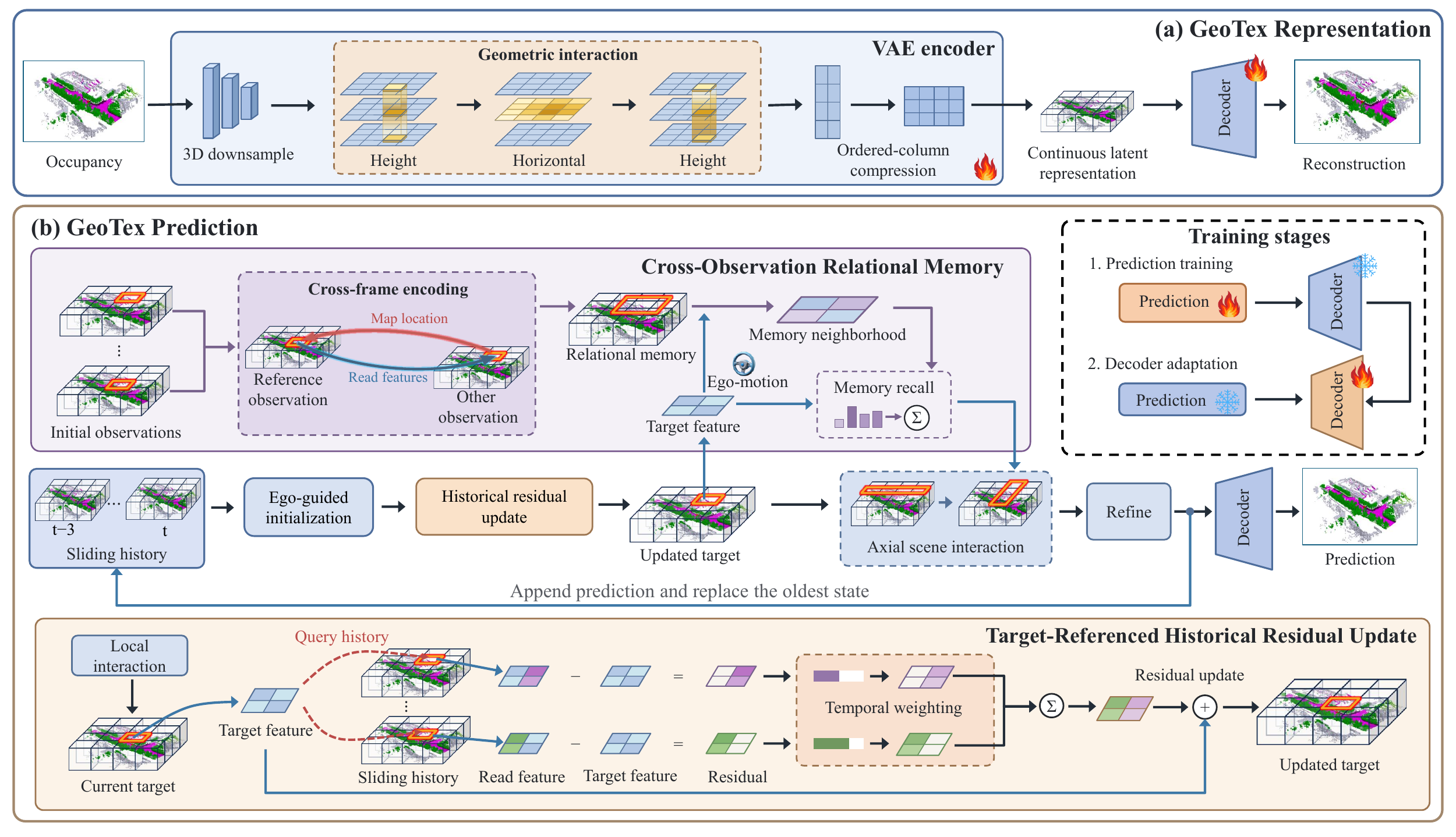}
    \caption{\textbf{Overview of GeoTex.} (a) GeoTex Representation organizes height patterns and cross-column relations into compact continuous latent representations for scene reconstruction. (b) GeoTex Prediction grounds autoregressive forecasting in cross-observation relational memory and target-referenced historical residual updates. Decoder-only adaptation further improves semantic decoding of predicted states.}
    \label{fig:framework}
\end{figure*}

\subsection{3D Scene Representation and Latent Encoding}
\label{sec:related_representation}

3D scene representation research organizes scene geometry through height information and spatial context. BEVDepth~\cite{bevdepth} and BEVHeight~\cite{bevheight} connect image features to 3D space using depth and ground-relative height, respectively, while 3D occupancy perception methods further recover scene structure through spatial context~\cite{monoscene,voxformer,occformer,panoocc}. To efficiently forecast scene evolution, occupancy world models draw on image and video tokenizers~\cite{vqgan,magvit2,titok,ivideogpt} to construct latent states through voxel encoding~\cite{renderworld}, temporal reconstruction~\cite{uniscene}, and compact planar representations~\cite{dynamiccity,occworld,i2world,geniedrive}. However, existing encoders insufficiently organize within-column height patterns and cross-column relations jointly, weakening local shape cues. GeoTex Representation jointly organizes height-wise and horizontal relations before compression and restores 3D structure through content-dependent height unfolding during decoding, preserving geometric texture throughout encoding and reconstruction.

\subsection{4D Occupancy Forecasting and World Models}
\label{sec:related_forecasting}

Occupancy world models primarily model future scenes through diffusion generation~\cite{occsora,dome,come} or autoregressive prediction~\cite{occworld,i2world,geniedrive}, while OccLLaMA~\cite{occllama} and Occ-LLM~\cite{occllm} further introduce language conditions to expand generation control. To exploit motion conditions and spatial structure, DFIT-OccWorld~\cite{dfitoccworld} and OccSim~\cite{occsim} introduce dynamic flow and rigid propagation, respectively, while EfficientOCF~\cite{efficientocf} and GAST~\cite{gast} combine geometric propagation with content refinement to improve future scene prediction. However, even with geometric correspondence, direct aggregation does not explicitly distinguish what each historical frame adds to the current state. GeoTex Prediction preserves initial observation evidence in fixed relational memory and extracts target-referenced residual updates from recent history, balancing reliable observation evidence with recent state changes across prediction steps.

\section{Proposed Method}
\label{sec:method}

\subsection{Overall Architecture}
\label{sec:overall}

As shown in \cref{fig:framework}, GeoTex consists of GeoTex Representation and GeoTex Prediction, combining geometry-as-texture encoding with history-grounded forecasting. Given $L$ historical occupancy frames $\mathbf O_{-L+1:0}$, GeoTex forecasts $T$ future semantic occupancy frames $\widehat{\mathbf O}_{1:T}$. Each frame is $\mathbf O_t\in\{0,\ldots,K-1\}^{H\times W\times D}$, where $H,W,D$ are the grid dimensions and $K$ is the number of labels. GeoTex Representation jointly organizes height patterns and cross-column relations, mapping each scene through the encoder to a continuous latent state $\mathbf Z_t=\operatorname{Enc}(\mathbf O_t)\in\mathbb R^{h\times w\times c}$, where $h,w$ are the latent planar resolutions and $c$ is the channel dimension. In this latent space, the prediction module $F$ combines ego-motion conditions, fixed relational memory built from initial observations, and recent sliding history to autoregressively generate future states $\widehat{\mathbf Z}_{1:T}$. The decoder $\operatorname{Dec}$ then reconstructs future occupancy from these states.

Training first learns GeoTex Representation through occupancy reconstruction to preserve geometric texture during compression and reconstruction; its encoder and decoder are detailed in \cref{sec:representation}. We then freeze the representation module and train GeoTex Prediction to model scene evolution conditioned on history in latent space. Finally, we freeze the encoder and prediction module and adapt only the decoder to improve semantic decoding of predicted states. The prediction module and decoder adaptation are described in \cref{sec:predictor}.

\subsection{GeoTex Representation}
\label{sec:representation}

Semantic occupancy explicitly describes 3D positions and spatial occupancy, but abstracts continuous appearance into discrete classes. Voxels with the same semantic label may play different structural roles, such as the interior, outer contour, and top termination of a tree canopy. These roles are revealed by voxel relations: vertical continuation, gaps, and termination characterize within-column shape, while changes and connections between neighboring columns describe surfaces, boundaries, and spatial extent. We refer to the discriminative cues jointly formed by these height patterns and their horizontal organization as \emph{geometric texture}.

To preserve these geometric relations in a compact representation, GeoTex Representation organizes 3D geometric features before height compression within a continuous VAE framework. Class embedding and local 3D encoding first produce $\mathbf S\in\mathbb R^{h\times w\times d\times C}$, where $d$ and $C$ denote the encoded height resolution and feature dimension. Attention blocks~\cite{transformer} then connect within-column shape with surrounding structures in a height--horizontal--height order:
\begin{equation}
\begin{aligned}
\mathbf S
&=\operatorname{Conv}^{3D}_{\downarrow}
  \big(\operatorname{Embed}(\mathbf O)\big),\\
\mathbf S^v_{u,:}
&=\operatorname{Attn}^{(1)}_z
  \big(\mathbf S_{u,:}\mid\mathbf e_z\big),\\
\mathbf S^h_{:,\ell}
&=\operatorname{Attn}_{xy}
  \big(\mathbf S^v_{:,\ell}\big),\\
\mathbf G_{u,:}
&=\operatorname{Attn}^{(2)}_z
  \big(\mathbf S^h_{u,:}\mid\mathbf e_z\big),
\end{aligned}
\label{eq:geotex_organization}
\end{equation}
where $u$ and $\ell$ index horizontal position and height, and $\mathbf e_z$ denotes ordered height embeddings learned separately in each block. The first height interaction produces column-aware features $\mathbf S^v$. Conditioned on these features and relative positions, horizontal interaction adaptively aggregates cross-column information at each height in parallel along the two planar axes to form $\mathbf S^h$. The second height interaction integrates the horizontal context across levels, linking column shape with surrounding geometry in $\mathbf G$ before compression. All stages use residual updates, with $\mathbf e_z$ added to height-attention queries and keys.

After these geometric interactions, we compress the volume features $\mathbf G$ into a planar representation. At each horizontal position $u$, ordered-column compression applies a learned mapping to $\mathbf G_{u,1}\oplus\cdots\oplus\mathbf G_{u,d}$, where $\oplus$ denotes concatenation in height order. Height-ordered concatenation keeps features from different levels distinguishable before aggregation, allowing structures with similar semantic compositions but different vertical arrangements to enter compression as distinct feature sequences. Planar refinement and posterior heads then parameterize the diagonal Gaussian posterior $q(\cdot\mid\mathbf O)$~\cite{kingma2014vae}, whose mean defines the latent representation $\mathbf Z=\operatorname{Enc}(\mathbf O)$.

To recover 3D structure from the compact latent state, the decoder produces semantic logits through $\widetilde{\mathbf O}=\operatorname{Dec}(\mathbf Z)$. It first extracts local planar context and unfolds volume features through height-specific content mappings:
\begin{equation}
\begin{aligned}
\mathbf V
&=\operatorname{Context}(\mathbf Z),\\
\widetilde{\mathbf G}_{u,\ell}
&=L_\ell(\mathbf V_u)+\mathbf e_{z,\ell},
\end{aligned}
\label{eq:ordered_codec}
\end{equation}
where $\mathbf V_u$ is the planar context at horizontal position $u$. The height-specific mapping $L_\ell$ and decoder height embedding $\mathbf e_{z,\ell}$ produce the unfolded volume feature $\widetilde{\mathbf G}_{u,\ell}$ at height level $\ell$. The decoder then applies height--horizontal--height interaction to the unfolded volume with its own parameters, followed by 3D upsampling to produce semantic logits.

We supervise this encoding and decoding process with the occupancy reconstruction loss
\begin{equation}
\mathcal L_{\rm occ}
=\mathcal L_{\rm CE}+\mathcal L_{\rm Lov},
\label{eq:occupancy_loss}
\end{equation}
where $\mathcal L_{\rm CE}$ and $\mathcal L_{\rm Lov}$ are cross-entropy and Lov\'asz--Softmax~\cite{lovasz} losses. For posterior samples $\mathbf z$, we combine this reconstruction loss with KL regularization:
\begin{equation}
\begin{aligned}
\mathcal L_{\rm repr}
&=\mathbb E_{\mathbf z\sim q(\cdot\mid\mathbf O)}
  \left[
  \mathcal L_{\rm occ}
  \big(\operatorname{Dec}(\mathbf z),\mathbf O\big)
  \right]\\
&\quad+\frac{\lambda_{\rm KL}}{hwc}
  D_{\rm KL}\!\left(
  q(\mathbf z\mid\mathbf O)
  \,\|\,
  \mathcal N(\mathbf 0,\mathbf I)
  \right),
\end{aligned}
\label{eq:representation_loss}
\end{equation}
where $\mathcal N(\mathbf 0,\mathbf I)$ is the standard Gaussian prior and $\lambda_{\rm KL}$ weights the mean KL divergence over $hwc$ latent elements. Forecasting uses the posterior mean $\mathbf Z$.

\subsection{GeoTex Prediction}
\label{sec:predictor}

In this continuous latent space, GeoTex Prediction learns how scene states evolve over time. History-grounded forecasting requires both a persistent reference to observed structure and complementary information about recent evolution. We therefore organize initial observations into fixed relational memory $\mathcal M$ and extract target-referenced residual updates from the sliding history $\mathcal H_t$. At autoregressive step $t$, the prediction module combines these two sources with ego-motion conditions to generate the next latent state:
\begin{equation}
\widehat{\mathbf Z}_{t+1}
=F(\mathcal H_t,\mathcal M,\mathbf a_t),
\qquad t=0,\ldots,T-1,
\label{eq:forecast}
\end{equation}
where $\mathcal H_t=\{(\bar{\mathbf Z}_s,\mathbf P_s)\}_{s=t-L+1}^{t}$ stores recent latent states and ego-to-global pose matrices $\mathbf P_s\in\mathbb R^{4\times4}$. During inference, $\bar{\mathbf Z}_s$ is an initially encoded state $\mathbf Z_s$ for $s\leq0$ or a previously predicted state $\widehat{\mathbf Z}_s$ for $s>0$. The condition $\mathbf a_t$ encodes ego-motion from $t$ to $t+1$. Ego-motion establishes spatial correspondence between target locations and historical content, while current target content guides feature retrieval and updates.

\paragraph{Cross-Observation Relational Memory.}
The value of history lies not only in what each frame observes, but also in how different observations describe the same scene. A local shape can receive complementary evidence from another viewpoint, while a boundary can obtain more complete spatial context through cross-frame relations. We take each initial observation in turn as the reference and read related content from the remaining frames, enhancing its representation with cross-frame context. Let $\mathcal I_0=\{-L+1,\ldots,0\}$ denote the initial frame index set, and $\mathbf T_{j\rightarrow i}=\mathbf P_i^{-1}\mathbf P_j$ map coordinates from source frame $j$ to reference frame $i$. Relational encoding is written as
\begin{equation}
\begin{aligned}
\mathbf R_i^{\rm mem}
&=\operatorname{CrossRead}\!\left(
\mathbf Z_i,
\{\mathbf Z_j,\mathbf T_{j\rightarrow i}\}_{j\in\mathcal I_0,\,j\ne i}
\right),\\
\mathbf B_i
&=\operatorname{Spatial}\!\left(
\mathbf Z_i+\mathbf R_i^{\rm mem}
\right),\\
\mathcal M
&=\{(\mathbf Z_i,\mathbf B_i,\mathbf P_i)\}_{i\in\mathcal I_0},
\end{aligned}
\label{eq:memory}
\end{equation}
where $\operatorname{CrossRead}$ retrieves neighborhood content $\mathbf R_i^{\rm mem}$, and $\operatorname{Spatial}$ organizes planar context. Memory retains native states $\mathbf Z_i$ and relational states $\mathbf B_i$. Relational features guide selection and supplement retrieved native content to form target-time scene evidence. The cross-frame encoding branch in \cref{fig:framework} constructs this memory once, keeping observed evidence accessible throughout the rollout.

\paragraph{Target-Referenced Historical Residual Update.}
Fixed relational memory preserves scene context from initial observations, while recent state changes enter prediction through the sliding history. We transport the latest state according to ego-motion into the next-frame coordinate system to initialize the target state, then apply local interaction and historical residual updates through successive transition units. Given a unit's input state $\mathbf U$, local interaction produces $\mathbf U_{\rm loc}=\operatorname{Local}(\mathbf U)$, which retrieves features $\mathbf R_s$ from each recent frame $s$ for the current target positions:
\begin{equation}
\mathbf R_s
\!=\!\operatorname{GeoAttn}\!\left(
\operatorname{Query}(\mathbf U_{\rm loc},\mathbf a_t),
\bar{\mathbf Z}_s;
\operatorname{Grid}(\mathbf T_{t+1\rightarrow s})
\right),
\label{eq:history_read}
\end{equation}
where $\operatorname{Grid}(\mathbf T_{t+1\rightarrow s})$ maps target locations into source frame $s$, and $\operatorname{Query}$ combines target content, position, and ego-motion. $\operatorname{GeoAttn}$ reads relevant content around these geometric references using learned sampling offsets~\cite{deformabledetr}.

To extract complementary information from each historical frame relative to the current target, we subtract projected target features from the retrieved features to form target-referenced residuals $\boldsymbol\delta_s$, then weight them across frames to obtain the historical update $\Delta\mathbf U_{\rm hist}$:
\begin{equation}
\begin{aligned}
\boldsymbol{\delta}_s
&=\mathbf R_s-\boldsymbol\nu_s\odot\big(\mathbf W_v\mathbf U_{\rm loc}\big),\\
\Delta\mathbf U_{\rm hist}
&=\operatorname{HistProj}\!\left(\sum_{s=t-L+1}^{t}\boldsymbol\beta_s\odot\boldsymbol\delta_s\right),
\end{aligned}
\label{eq:historical_residual}
\end{equation}
where $\mathbf W_v$ denotes the channel projection shared by historical and target features within each transition unit, and $\mathbf R_s$ aggregates the projected historical features. This provides a common feature space and target reference for residuals across frames. The mask $\boldsymbol\nu_s$ restricts target-feature subtraction to supported reads; $\odot$ denotes element-wise multiplication. Ordered residuals and their adjacent changes determine $\boldsymbol\beta_s$, normalized over valid historical frames at each target position and broadcast over channels. $\operatorname{HistProj}$ then maps the weighted residual sum to the target feature space.

\paragraph{History-Grounded Forecasting.}
After historical residual updating, the memory-enhanced transition unit uses $\mathbf U_h$ to query fixed relational memory. Axial scene interaction then integrates recalled evidence with target content to form $\mathbf U_g$:
\begin{equation}
\begin{aligned}
\mathbf U_h
&=\mathbf U_{\rm loc}+\Delta\mathbf U_{\rm hist},\\
(\mathbf R^{\mathrm{recall}},\mathbf m)
&=\operatorname{Recall}(\mathbf U_h,\mathcal M,\mathbf a_t),\\
\mathbf U_g
&=\operatorname{SceneAttn}(\mathbf U_h,\mathbf R^{\mathrm{recall}},\mathbf m,\mathbf a_t),\\
\widehat{\mathbf Z}_{t+1}
&=\operatorname{Refine}(\mathbf U_g,\mathcal H_t,\mathbf a_t),
\end{aligned}
\label{eq:local_global_loop}
\end{equation}
where $\mathbf R^{\mathrm{recall}}$ and $\mathbf m$ are recalled evidence on the target grid and its validity mask. $\operatorname{SceneAttn}$ applies axial attention~\cite{axialdeeplab} along both planar axes. $\operatorname{Refine}$ continues local spatial interaction and historical updates to form the next latent state, which is appended to the sliding history.

To learn these state transitions, we freeze GeoTex Representation and train the prediction module and memory encoder with future-latent mean squared error (MSE):
\begin{equation}
\mathcal L_{\rm pred}
=\sum_{\tau=1}^{T}w_\tau\,
\operatorname{MSE}\!\left(
\widehat{\mathbf Z}_\tau,
\operatorname{Enc}(\mathbf O_\tau)
\right),
\label{eq:prediction_loss}
\end{equation}
where $\tau$ indexes future frames, $w_\tau$ weights each forecast step, and MSE averages over latent elements. We reduce teacher forcing through scheduled sampling~\cite{scheduledsampling} and retain cross-step gradients through generated states.

\paragraph{Prediction-Domain Decoder Adaptation.}
Following this training, the prediction module generates future latent states, which the decoder restores to semantic occupancy. Since representation learning exposes the decoder only to states encoded from ground-truth occupancy, we then freeze the encoder and prediction module and adapt only the decoder using both encoded and predicted states to improve semantic decoding of generated states. For future frame $\tau$, the encoded state $\mathbf Z_\tau=\operatorname{Enc}(\mathbf O_\tau)$ and predicted state $\widehat{\mathbf Z}_\tau$ share the same decoder:
\begin{equation}
\begin{aligned}
\mathcal L_{\rm adapt}
=\mathbb E_\tau\Big[
&\eta\,
\mathcal L_{\rm occ}
\big(\operatorname{Dec}(\mathbf Z_\tau),\mathbf O_\tau\big)\\
&+(1-\eta)\,
\mathcal L_{\rm occ}
\big(\operatorname{Dec}(\widehat{\mathbf Z}_\tau),\mathbf O_\tau\big)
\Big],
\end{aligned}
\label{eq:adaptation_loss}
\end{equation}
where $\mathbb E_\tau$ averages over future steps and $\eta$ balances the encoded and predicted domains. The first term retains reconstruction supervision, while the second adapts semantic readout to predicted states.

\section{Experiments}
\label{sec:exp}

\begin{table*}[t]
  \centering
  \caption{\textbf{4D occupancy forecasting on Occ3D-nuScenes.} The upper block reports endpoint results, where Avg. is the arithmetic mean over 1s, 2s, and 3s. Following the public GAST evaluator, the lower block pools adjacent future-frame pairs for the three intervals and all six frames for Avg. Best and second-best upper-block accuracy results are in \textbf{bold} and \underline{underlined}, respectively.}
  \label{tab:main_forecasting}
  \setlength{\tabcolsep}{2.8pt}
  \resizebox{\textwidth}{!}{%
  \begin{tabular}{l|c|ccccc|ccccc|cc}
    \toprule
    \multirow{2}{*}{Method} & \multirow{2}{*}{Input} & \multicolumn{5}{c|}{mIoU (\%)$\uparrow$} & \multicolumn{5}{c|}{IoU (\%)$\uparrow$} & \multirow{2}{*}{FPS$\uparrow$} & \multirow{2}{*}{Params. (M)$\downarrow$} \\
    \cmidrule(lr){3-7}\cmidrule(lr){8-12}
    & & Recon. & 1s & 2s & 3s & Avg. & Recon. & 1s & 2s & 3s & Avg. & & \\
    \midrule
    OccWorld~\cite{occworld} & Occ & 66.38 & 25.78 & 15.14 & 10.51 & 17.14 & 62.29 & 34.63 & 25.07 & 20.18 & 26.63 & 18.00 & 72.39 \\
    OccSora~\cite{occsora} & Occ & 66.97 & 32.77 & 22.04 & 14.40 & 23.07 & 68.78 & 41.39 & 33.68 & 29.97 & 35.01 & 20.00 & 174.19 \\
    OccLLaMA~\cite{occllama} & Occ \& Text & 75.20 & 25.05 & 19.49 & 15.26 & 19.93 & 63.79 & 34.56 & 28.53 & 24.41 & 29.17 & -- & $>$7000 \\
    Occ-LLM~\cite{occllm} & Occ \& Text & -- & 24.02 & 21.65 & 17.29 & 20.99 & -- & 36.65 & 32.14 & 28.77 & 32.52 & -- & $>$7000 \\
    DFIT~\cite{dfitoccworld} & Occ \& Camera & -- & 31.68 & 21.29 & 15.19 & 22.71 & -- & 40.28 & 31.24 & 25.29 & 32.27 & -- & -- \\
    DOME~\cite{dome} & Occ & 83.08 & 35.11 & 25.89 & 20.29 & 27.10 & \underline{77.25} & 43.99 & 35.36 & 29.74 & 36.36 & 6.54 & 397.55 \\
    UniScene~\cite{uniscene} & Occ \& Box \& Map & 72.90 & 35.37 & 29.59 & 25.08 & 30.01 & 64.10 & 38.34 & 32.70 & 29.09 & 33.38 & 1.72 & 69.47 \\
    COME~\cite{come} & Occ & 83.08 & 42.75 & 32.97 & 26.98 & 34.23 & \underline{77.25} & 50.57 & 43.47 & 38.36 & 44.13 & 0.30 & 692.97 \\
    I$^2$-World~\cite{i2world} & Occ & 81.22 & 47.62 & 38.58 & 32.98 & 39.73 & 68.30 & 54.29 & 49.43 & 45.69 & 49.80 & 37.04 & 22.71 \\
    GenieDrive~\cite{geniedrive} & Occ & \underline{86.15} & \underline{50.47} & \underline{41.47} & \underline{35.83} & \underline{42.59} & 75.53 & \underline{56.87} & \underline{51.46} & \underline{47.08} & \underline{51.80} & 41.38 & 3.47 \\
    \rowcolor{black!10}
    \textbf{GeoTex(Ours)} & Occ & \textbf{95.03} & \textbf{54.54} & \textbf{45.99} & \textbf{39.76} & \textbf{46.76} & \textbf{88.40} & \textbf{62.59} & \textbf{57.79} & \textbf{53.54} & \textbf{57.97} & 47.28 & \textbf{2.33} \\
    \midrule
    \multicolumn{14}{l}{\textit{Adjacent-pair pooled protocol (GAST evaluator)}} \\
    GAST~\cite{gast} & Occ & -- & 55.54 & 46.33 & 40.18 & 47.40 & -- & 60.77 & 55.97 & 51.96 & 56.24 & -- & 12.09 \\
    \rowcolor{black!10}
    \textbf{GeoTex(Ours)} & Occ & -- & \textbf{58.03} & \textbf{47.69} & \textbf{41.27} & \textbf{48.93} & -- & \textbf{64.54} & \textbf{58.89} & \textbf{54.58} & \textbf{59.34} & 47.28 & \textbf{2.33} \\
    \bottomrule
  \end{tabular}
  }
\end{table*}

\subsection{Experimental Settings}

\paragraph{Dataset and Metrics.}
Following prior occupancy world models~\cite{occworld,i2world,geniedrive,gast}, we conduct experiments on Occ3D-nuScenes~\cite{occ3d}. Built on nuScenes~\cite{nuscenes}, it contains 700 training and 150 validation scenes annotated at 2\,Hz. Each frame is a $200\!\times\!200\!\times\!16$ grid spanning $[-40,40]$ meters horizontally and $[-1,5.4]$ meters vertically at 0.4-meter resolution. Four historical occupancy frames are used to predict six future frames. Reconstruction and forecasting are evaluated by mIoU over 17 non-empty classes and IoU over their union. Short-term results use 1s, 2s, and 3s endpoints and their arithmetic mean. Long-term results use 4s, 5s, and 6s.

\paragraph{Implementation Details.}
GeoTex is implemented in PyTorch and trained on two NVIDIA RTX 4090 GPUs. GeoTex Representation and GeoTex Prediction use AdamW~\cite{adamw} with learning rates of $3\!\times\!10^{-4}$ and $10^{-3}$, respectively. We then adapt only the decoder for 10 epochs at $10^{-5}$. Architecture and additional training details are provided in Appendix~\ref{sec:supp_hyperparameters}.

\subsection{Evaluation on 4D Occupancy Forecasting}
\label{sec:forecasting_evaluation}

\paragraph{Short-term Occupancy Forecasting.}
As shown in \cref{tab:main_forecasting}, GeoTex achieves the best mIoU and IoU at 1s, 2s, and 3s. Average mIoU increases from GenieDrive's 42.59\% to 46.76\%, a relative gain of 9.79\%. Under GAST's adjacent-frame pooled protocol, GeoTex also achieves relative gains of 3.23\% and 5.51\% over GAST in average mIoU and IoU, respectively.

\begin{figure*}[t]
  \centering
  \includegraphics[width=0.96\textwidth]{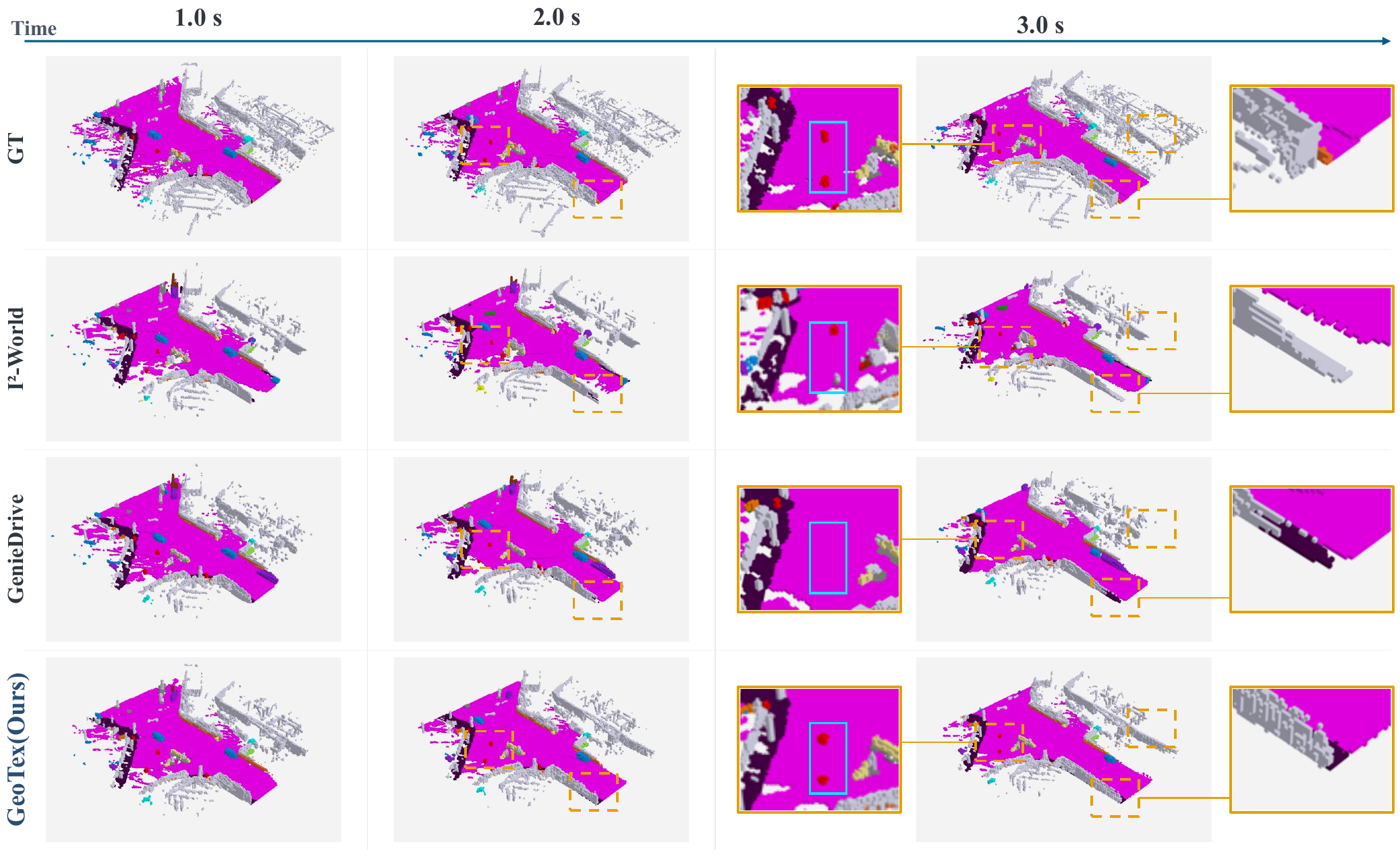}
  \caption{\textbf{Qualitative comparison of short-term occupancy forecasting on Occ3D-nuScenes.} Rows show the ground truth and predictions from I$^2$-World, GenieDrive, and GeoTex at 1s, 2s, and 3s. The enlarged regions at 3s highlight pedestrian prediction (left, cyan boxes) and building-structure continuation (right).}
  \label{fig:short_term_qualitative}
\end{figure*}

GeoTex reaches 57.97\% average IoU, an 11.91\% relative gain over GenieDrive, exceeding the relative mIoU gain. Since IoU measures occupancy overlap independently of semantic labels, this improvement reflects better geometric prediction. At 3s, GeoTex more accurately predicts pedestrians in the left inset's cyan boxes and preserves wall continuity in the right inset, whereas baselines miss foreground details or fragment geometry (\cref{fig:short_term_qualitative}).

Alongside these accuracy gains, GeoTex uses only 2.33M parameters, while GeoTex Prediction achieves an inference speed of 47.28 FPS. Observation memory is encoded once per rollout and reused alongside a four-frame sliding history, retaining initial evidence without enlarging the history window. Compared with GenieDrive's $50\!\times\!58$ latent grid, GeoTex's $50\!\times\!50$ grid further reduces per-location projection and feature-update costs.

\begin{table}[t]
  \centering
  \caption{\textbf{Long-term 4D occupancy forecasting on Occ3D-nuScenes.} Avg. is the mean over 4s, 5s, and 6s. GAST values are reproduced as published.}
  \label{tab:long_term_forecasting}
  \setlength{\tabcolsep}{3.1pt}
  \resizebox{\columnwidth}{!}{%
  \begin{tabular}{@{}l|cccc|cccc@{}}
    \toprule
    \multirow{2}{*}{Method} & \multicolumn{4}{c|}{mIoU (\%)$\uparrow$} & \multicolumn{4}{c}{IoU (\%)$\uparrow$} \\
    \cmidrule(lr){2-5}\cmidrule(lr){6-9}
    & 4s & 5s & 6s & Avg. & 4s & 5s & 6s & Avg. \\
    \midrule
    OccWorld~\cite{occworld} & 7.79 & 6.29 & 5.76 & 6.61 & 17.23 & 15.31 & 14.54 & 15.69 \\
    DOME~\cite{dome} & 23.50 & 19.37 & 17.27 & 20.05 & 13.84 & 10.14 & 8.15 & 10.71 \\
    UniScene~\cite{uniscene} & 16.44 & 16.41 & 15.61 & 16.15 & 14.31 & 14.33 & 14.15 & 14.26 \\
    I$^2$-World~\cite{i2world} & 28.59 & 25.32 & 22.55 & 25.49 & 41.75 & 38.47 & 35.41 & 38.54 \\
    GenieDrive~\cite{geniedrive} & 31.16 & 27.17 & 23.66 & 27.33 & 42.81 & 39.00 & 35.60 & 39.14 \\
    GAST~\cite{gast} & 24.36 & 21.98 & 19.81 & 22.05 & 39.52 & 37.19 & 34.99 & 37.23 \\
    \midrule
    \textbf{GeoTex(Ours)} & \textbf{35.83} & \textbf{32.14} & \textbf{28.86} & \textbf{32.28} & \textbf{49.94} & \textbf{46.61} & \textbf{43.35} & \textbf{46.63} \\
    \bottomrule
  \end{tabular}
  }
\end{table}

\paragraph{Long-term Occupancy Forecasting.}
GeoTex achieves the best reported results from 4s to 6s, averaging 32.28\% mIoU and 46.63\% IoU (\cref{tab:long_term_forecasting}). Compared with GenieDrive, relative gains increase from 15.0\% at 4s to 22.0\% at 6s in mIoU and from 16.7\% to 21.8\% in IoU, showing a widening advantage at longer horizons.

The widening advantage reflects slower forecasting degradation. From 4s to 6s, GeoTex's mIoU and IoU decrease by 19.45\% and 13.20\%, compared with 24.07\% and 16.84\% for GenieDrive. This slower decline highlights GeoTex's stability over multiple autoregressive steps. GeoTex Representation preserves a stable geometric basis for long-term forecasting, while persistent observation memory enables GeoTex Prediction to retrieve reliable scene evidence after initial observations leave the sliding window, mitigating structural degradation and forecasting drift.

\subsection{Geometry-Preserving Scene Reconstruction}
\label{sec:representation_evaluation}

We compare GeoTex Representation with I$^2$-World and GenieDrive on the complete validation set, assessing reconstruction quality and fidelity at interfaces and interiors.

\begin{table}[t]
  \centering
  \caption{\textbf{Scene reconstruction.} Interfaces are occupied voxels with an empty or differently labeled GT six-neighbor. Other occupied voxels are interiors.}
  \label{tab:representation_reconstruction}
  \setlength{\tabcolsep}{2.8pt}
  \resizebox{\columnwidth}{!}{%
  \begin{tabular}{@{}llcccc@{}}
    \toprule
    \multirow{3}{*}{Method} & \multirow{3}{*}{Encoding} & \multicolumn{4}{c}{Reconstruction} \\
    \cmidrule(lr){3-6}
    & & \multirow{2}{*}{mIoU (\%)$\uparrow$} & \multirow{2}{*}{IoU (\%)$\uparrow$} & Interface & Interior \\
    & & & & mIoU (\%)$\uparrow$ & mIoU (\%)$\uparrow$ \\
    \midrule
    I$^2$-World~\cite{i2world} & Intra--Inter Tokens & 81.22 & 68.30 & 86.88 & 97.34 \\
    GenieDrive~\cite{geniedrive} & Tri-plane & 86.15 & 75.53 & 91.18 & 98.64 \\
    \textbf{GeoTex(Ours)} & \textbf{Geometry-as-Texture} & \textbf{95.03} & \textbf{88.40} & \textbf{96.71} & \textbf{99.75} \\
    \bottomrule
  \end{tabular}
  }
\end{table}

\begin{figure}[t]
  \centering
  \begingroup
  \setlength{\unitlength}{\columnwidth}
  \begin{picture}(1,0.423611)
    \put(0,0.049383){\includegraphics[width=\unitlength,trim=0 25.6bp 0 0,clip]{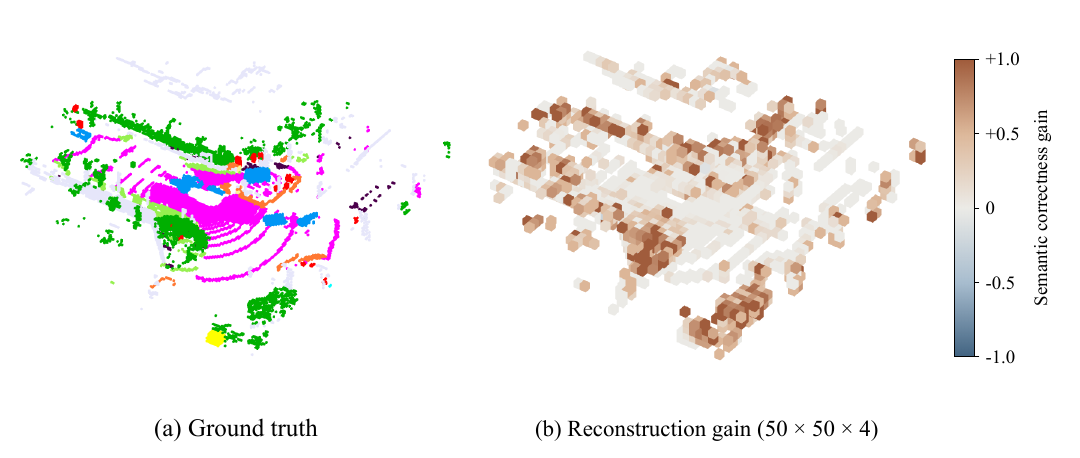}}
    \put(0.2185,0.0243){\makebox(0,0){\paneltitle{(a) Ground truth}}}
    \put(0.6544,0.0243){\makebox(0,0){\paneltitle{(b) Reconstruction gain}}}
  \end{picture}
  \endgroup
  \caption{\textbf{Spatial distribution of reconstruction gains.} GeoTex's semantic-correctness gain over the mean of I$^2$-World and GenieDrive is aggregated into a $50\!\times\!50\!\times\!4$ field. Warm and cool colors denote improvements and regressions, respectively.}
  \label{fig:representation_spatial_gain}
\end{figure}

GeoTex Representation achieves the best performance across metrics in \cref{tab:representation_reconstruction}. Compared with GenieDrive, reconstruction mIoU increases from 86.15\% to 95.03\%, while IoU and interface mIoU improve by 17.04\% and 6.06\% relatively. All methods approach saturation in structural interiors, whereas interfaces retain a clear performance gap. The larger performance gap at structural interfaces highlights GeoTex's improved preservation of scene shape and semantic boundaries. These gains concentrate on regions with complex height changes, thin structures, and extended spatial layouts (\cref{fig:representation_spatial_gain}). Category-wise gains peak in manmade and vegetation, characterized by complex height variations and cross-column extent (Appendix~\ref{sec:supp_reconstruction_analysis}).

\subsection{Ablation Study}
\label{sec:ablation}

\paragraph{GeoTex Representation Evaluation.}
We construct two ablations to examine explicit geometric modeling and joint geometric interaction. The first merges height directly into channels and uses a 2D encoder and decoder, removing explicit 3D geometric modeling. The second retains local 3D feature extraction but removes the joint interaction between within-column height patterns and cross-column spatial relations. Both variants modify the decoder and use the same GeoTex Prediction architecture for forecasting.

\begin{table}[t]
  \centering
  \caption{\textbf{Representation ablations in reconstruction and forecasting.} Forecasting metrics average the 1s, 2s, and 3s endpoints after decoder adaptation.}
  \label{tab:representation_ablation}
  \setlength{\tabcolsep}{3pt}
  \resizebox{\columnwidth}{!}{%
  \begin{tabular}{@{}l|cccc|cc@{}}
    \toprule
    \multirow{3}{*}{Method} & \multicolumn{4}{c|}{Reconstruction (\%)} & \multicolumn{2}{c}{Forecasting (\%)} \\
    \cmidrule(lr){2-5}\cmidrule(lr){6-7}
    & \multirow{2}{*}{mIoU$\uparrow$} & \multirow{2}{*}{IoU$\uparrow$} & Interface & Interior & \multirow{2}{*}{mIoU$\uparrow$} & \multirow{2}{*}{IoU$\uparrow$} \\
    & & & mIoU$\uparrow$ & mIoU$\uparrow$ & & \\
    \midrule
    w/o Explicit Geometry Modeling & 84.13 & 76.19 & 89.63 & 99.16 & 43.57 & 53.45 \\
    w/o Joint Geometric Interaction & 91.82 & 85.32 & 94.80 & 99.51 & 44.46 & 55.51 \\
    \midrule
    \textbf{Full Method} & \textbf{95.03} & \textbf{88.40} & \textbf{96.71} & \textbf{99.75} & \textbf{46.76} & \textbf{57.97} \\
    \bottomrule
  \end{tabular}
  }
\end{table}

As shown in \cref{tab:representation_ablation}, retaining local 3D processing yields relative increases of 9.14\% in reconstruction mIoU and 11.99\% in IoU. This indicates that explicitly organizing 3D neighborhoods preserves scene structure more effectively than directly mixing height channels. Average forecasting mIoU and IoU also rise by 2.04\% and 3.87\%.

Full Method further connects within-column and cross-column structures, reaching 95.03\% reconstruction mIoU and 88.40\% IoU. Average forecasting scores reach 46.76\% and 57.97\%, relative gains of 5.17\% and 4.43\% over the second variant. All three variants approach saturation in interiors but differ markedly at structural interfaces. This aligns with the goal of joint geometric interaction: linking local height patterns with surrounding structure to preserve the geometric texture formed by interfaces, gaps, and height variations during compression and reconstruction.

\paragraph{GeoTex Prediction Component Evaluation.}
\Cref{tab:predictor_ablation} evaluates cross-observation relational memory, target-referenced historical residual update, and prediction-domain decoder adaptation.

\begin{table}[t]
  \centering
  \caption{\textbf{Component ablations of GeoTex Prediction on Occ3D-nuScenes.} Structural variants use adapted decoders. The w/o Decoder Adaptation variant pairs the final prediction module with the original decoder.}
  \label{tab:predictor_ablation}
  \setlength{\tabcolsep}{3.1pt}
  \resizebox{\columnwidth}{!}{%
  \begin{tabular}{@{}l|cccc|cccc@{}}
    \toprule
    \multirow{2}{*}{Method} & \multicolumn{4}{c|}{mIoU (\%)$\uparrow$} & \multicolumn{4}{c}{IoU (\%)$\uparrow$} \\
    \cmidrule(lr){2-5}\cmidrule(lr){6-9}
    & 1s & 2s & 3s & Avg. & 1s & 2s & 3s & Avg. \\
    \midrule
    w/o Relational Memory & 52.00 & 43.47 & 37.34 & 44.27 & 60.65 & 55.46 & 51.27 & 55.79 \\
    w/o Historical Residual Update & 50.68 & 42.68 & 36.53 & 43.30 & 60.36 & 55.33 & 51.05 & 55.58 \\
    w/o Decoder Adaptation & 47.72 & 38.87 & 32.85 & 39.81 & 60.72 & 55.89 & 51.61 & 56.07 \\
    \midrule
    \textbf{Full Method} & \textbf{54.54} & \textbf{45.99} & \textbf{39.76} & \textbf{46.76} & \textbf{62.59} & \textbf{57.79} & \textbf{53.54} & \textbf{57.97} \\
    \bottomrule
  \end{tabular}
  }
\end{table}

The memory-free variant trails Full Method by 5.33\% in average mIoU and 3.76\% in IoU, with the relative mIoU gap widening from 4.66\% at 1s to 6.09\% at 3s. Without historical residual updates, accuracy drops at every horizon, and average mIoU and IoU decline by 7.40\% and 4.12\%. Initial observation memory thus helps sustain multi-step accuracy, while complementary recent-history residuals support each state update.

Decoding Full Method's predictions with the original reconstruction decoder reduces average mIoU and IoU by 14.86\% and 3.28\%. The larger semantic decline reflects this decoder's limited ability to interpret predicted latent states, rather than a complete loss of scene structure in those states.

\begin{figure}[t]
  \centering
  \begin{subfigure}[t]{0.49\linewidth}
    \centering
    \includegraphics[width=\linewidth]{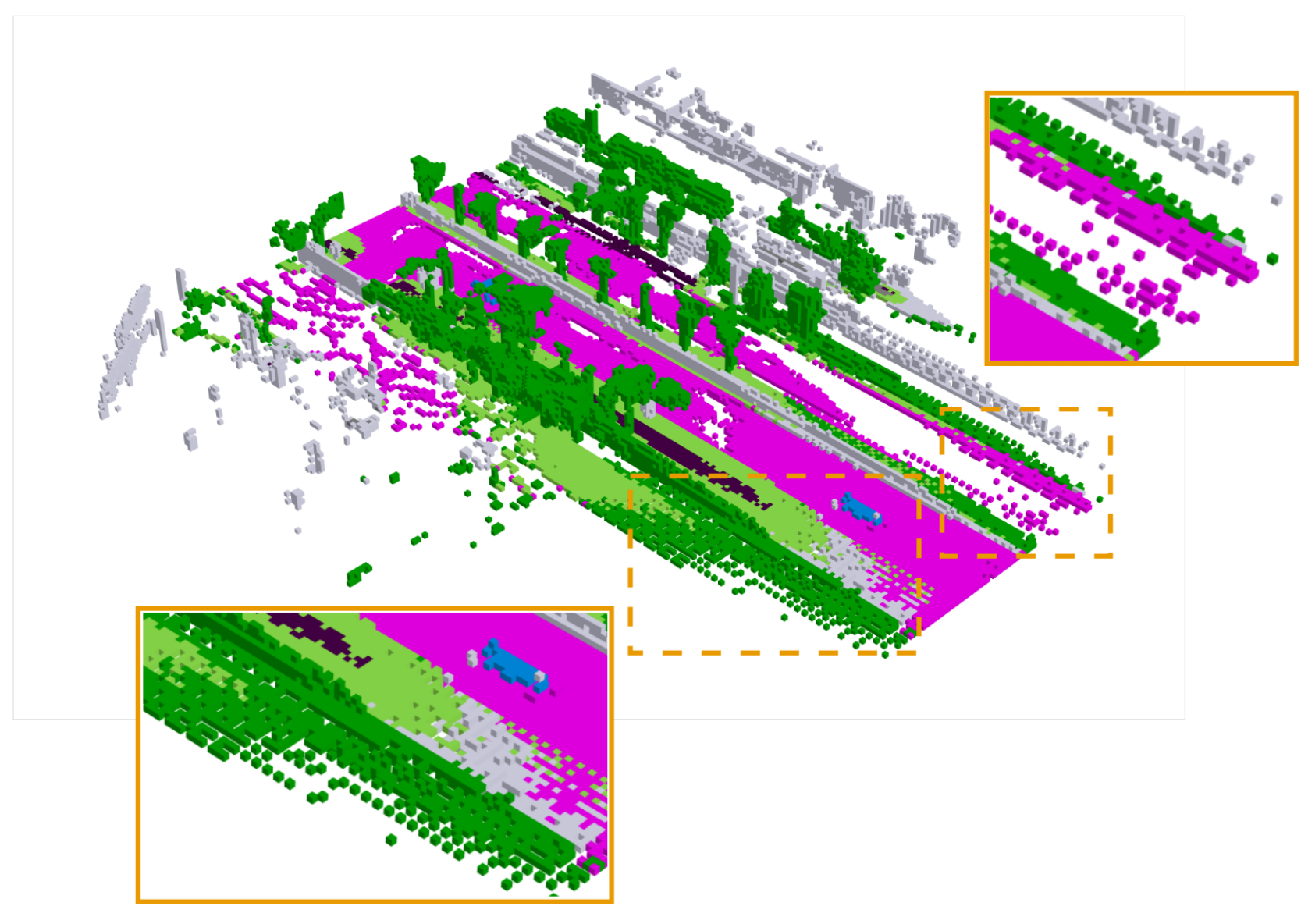}
    \caption{Before decoder adaptation}
    \label{fig:decoder_adaptation_before}
  \end{subfigure}\hfill
  \begin{subfigure}[t]{0.49\linewidth}
    \centering
    \includegraphics[width=\linewidth]{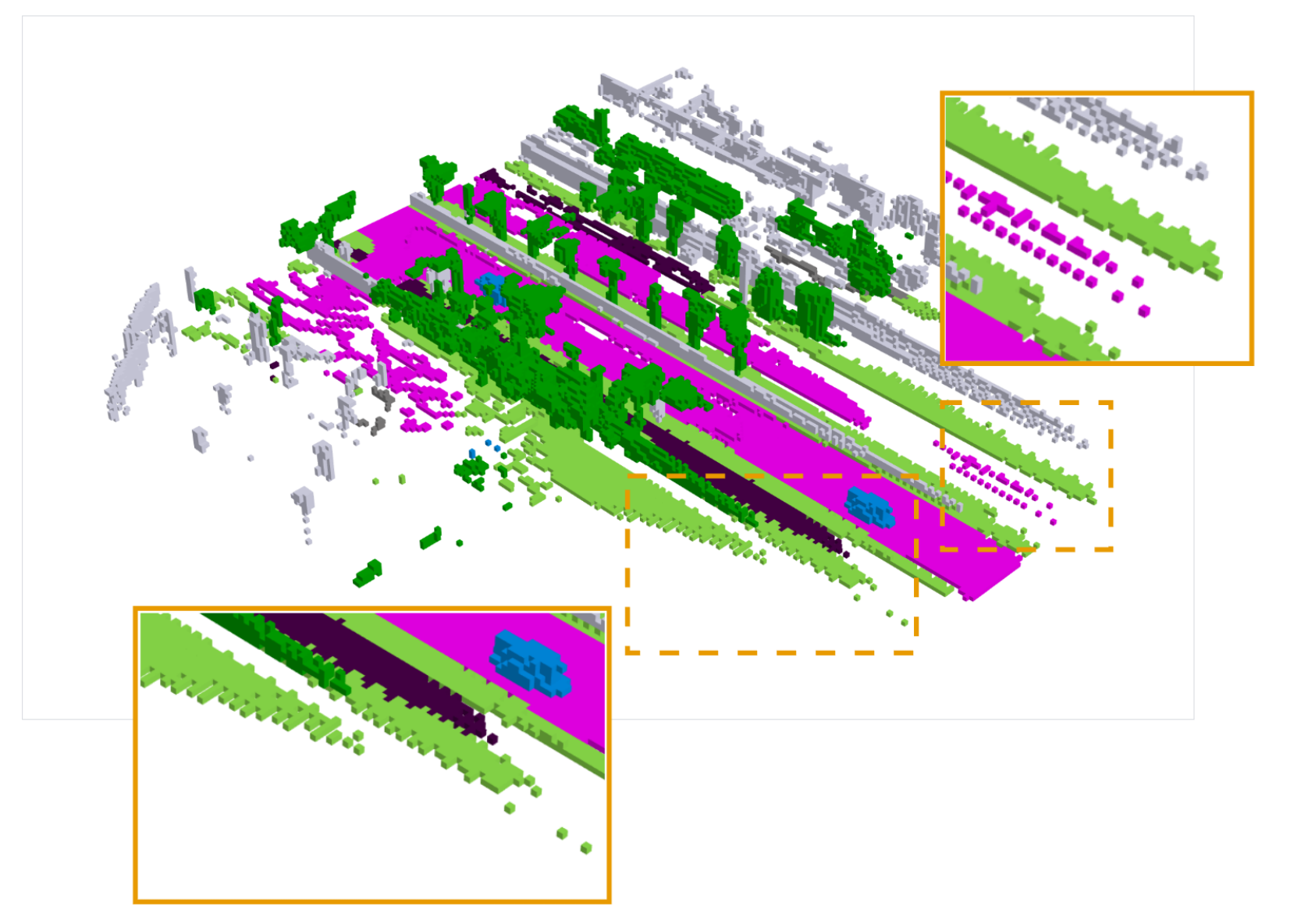}
    \caption{After decoder adaptation}
    \label{fig:decoder_adaptation_after}
  \end{subfigure}
  \caption{\textbf{Qualitative effect of prediction-domain decoder adaptation.} The enlarged regions highlight the grid-like semantic confusion and fragmented boundaries produced by the original reconstruction decoder, which are notably suppressed after adapting only the decoder to predicted latent states.}
  \label{fig:decoder_adaptation_qualitative}
\end{figure}

\paragraph{Prediction-Domain Decoder Adaptation.}
\label{sec:decoder_adaptation}
The decoder is trained on latent representations from observed occupancy but must decode autoregressively generated states during forecasting. This distribution shift can amplify inconsistencies between neighboring cells during height unfolding and two-stage 3D upsampling, producing grid-like semantic confusion and fragmented boundaries (\cref{fig:decoder_adaptation_qualitative}(a)).

With the encoder and prediction module frozen, we adapt the decoder using encoded and predicted states. Grid-like confusion decreases, and elongated structures and occupancy boundaries become more continuous (\cref{fig:decoder_adaptation_qualitative}(b)). Since both visualizations use identical predicted latent states, these improvements reflect better decoding of existing scene information. GeoTex's high pre-adaptation IoU and larger relative gains in mIoU than in IoU indicate that adaptation improves semantic decoding of geometry retained in predicted states.

Decoder adaptation also alleviates semantic confusion and merging of neighboring classes in I$^2$-World and GenieDrive. I$^2$-World's gains contrast with its E2E degradation reported by GenieDrive, highlighting the value of preserving the learned representation and prediction process. See Appendix~\ref{sec:supp_decoder_adaptation} for detailed comparisons.

\section{Conclusion}
\label{sec:conclusion}

We present GeoTex, an occupancy world model combining geometry-as-texture encoding with history-grounded 4D occupancy forecasting. GeoTex Representation jointly organizes height patterns and cross-column relations as geometric texture during compression and reconstruction. This preserves local shape distinctions within semantic classes. GeoTex Prediction retains initial observation evidence in fixed relational memory and updates future states through target-referenced historical residuals. Decoder-only adaptation then improves the decoding of these states into 3D semantic occupancy. Experiments demonstrate improved scene geometry and state-of-the-art forecasting with a compact model, and the gains increase at longer horizons.

\section*{LLM Usage Statement}
Generative AI tools, including OpenAI ChatGPT and Codex, were used during the preparation of this work to assist with language editing, improving the clarity and organization of the manuscript, and literature search and summarization. These tools also assisted with code syntax refinement and debugging, figure and table preparation and layout in LaTeX, and feedback on the presentation of mathematical formulations. All AI-generated suggestions were carefully assessed by the authors before incorporation. Relevant literature suggested by AI tools was manually verified against the original sources, and all AI-assisted text and technical content were reviewed and revised by the authors. The authors take full responsibility for the final content of this work, including all text, claims, results, and artifacts produced with the aid of generative AI.

{\small
\bibliographystyle{ieeenat_fullname}
\bibliography{references}
}
\clearpage
\appendix
\setcounter{figure}{0}
\setcounter{table}{0}
\setcounter{equation}{0}
\renewcommand{\thefigure}{A\arabic{figure}}
\renewcommand{\thetable}{A\arabic{table}}
\renewcommand{\theequation}{A\arabic{equation}}
\renewcommand{\theHfigure}{appendix.\arabic{figure}}
\renewcommand{\theHtable}{appendix.\arabic{table}}
\renewcommand{\theHequation}{appendix.\arabic{equation}}
\twocolumn[{\centering\Large\bfseries Appendix\par\vspace{1em}}]
\section{Architecture and Training Details}
\label{sec:supp_hyperparameters}

\begin{table}[H]
  \centering
  \caption{\textbf{Architecture hyperparameters of GeoTex.} The default $T=6$ corresponds to a 3-second forecasting horizon; $T=12$ is used only for long-term evaluation.}
  \label{tab:architecture_hyperparameters}
  \footnotesize
  \setlength{\tabcolsep}{2.6pt}
  \renewcommand{\arraystretch}{1.03}
  \begin{tabular}{@{}l|c|>{\raggedright\arraybackslash}p{0.52\columnwidth}@{}}
    \toprule
    Hyperparameter & Value & Description \\
    \midrule
    $H,W,D$ & $200,200,16$ & Dimensions of the semantic occupancy grid \\
    $K$ & $18$ & Semantic labels, including empty \\
    $L$ & $4$ & Number of historical observation frames \\
    $T$ & $6$ & Number of predicted future frames \\
    \midrule
    \multicolumn{3}{@{}l}{\textit{GeoTex Representation}} \\
    $C_{\mathrm{emb}}$ & $8$ & Semantic embedding dimension \\
    $C_{\mathrm{stem}},C$ & $48,96$ & 3D stem and feature channels \\
    $h,w,d$ & $50,50,4$ & Volume before column compression \\
    $N_{\mathrm{head}}^{\mathrm{rep}}$ & $4$ & Geometric-texture attention heads \\
    $h,w,c$ & $50,50,64$ & Continuous latent dimensions \\
    $C_{\mathrm{dec}}$ & $96,48,24$ & Decoder channel progression \\
    $\mathcal D_{\mathrm{local}}$ & $\{1,2\}$ & Local-context dilations \\
    \midrule
    \multicolumn{3}{@{}l}{\textit{GeoTex Prediction}} \\
    $|\mathcal M|$ & $4$ & Fixed observation-memory capacity \\
    $|\mathcal H_t|$ & $4$ & Sliding latent-history capacity \\
    $N_{\mathrm{cell}}$ & $5$ & Latent transition cells \\
    $i_{\mathrm{scene}}$ & $3$ & Full-scene interaction position \\
    $N_{\mathrm{head}}^{\mathrm{pred}}$ & $8$ & Prediction attention heads \\
    $N_{\mathrm{sample}}$ & $4$ & Samples per historical state \\
    -- & Axial $x/y$ & Factorization of full-scene interaction \\
    $C_{\mathrm{ego}}$ & $12\!\rightarrow\!64$ & Ego-motion projection \\
    \bottomrule
  \end{tabular}
\end{table}

Both representation learning and prediction training use a weight decay of $10^{-2}$.

\paragraph{Representation Learning.}
GeoTex Representation uses a learning rate of $3\!\times\!10^{-4}$. After 200 warm-up updates, cosine decay reduces the learning rate to $2\!\times\!10^{-6}$. The KL weight reaches $5\!\times\!10^{-3}$ over the first 2,000 updates.

\paragraph{Prediction Training.}
GeoTex Prediction uses a learning rate of $10^{-3}$, gradient clipping at 5, and an exponential moving average (EMA) with decay of 0.999.

\paragraph{Decoder Adaptation.}
We freeze the encoder and prediction module and adapt only the decoder for 10 epochs at a learning rate of $10^{-5}$, weighting the occupancy losses from encoded and predicted latent states equally.

\section{GeoTex Prediction Architecture}
\label{sec:supp_predictor}

GeoTex Prediction implements history-grounded forecasting through five transition cells. Each cell combines local spatial interaction with target-referenced historical residual updates from recent states. The third cell additionally recalls fixed observation memory and integrates this evidence through axially factorized full-scene interaction.

Memory recall supplies local evidence for each target position, but elongated structures, road extent, and cross-region dependencies require broader scene context. The third transition cell therefore updates the target state using both current content and recalled evidence through row-wise and column-wise interaction. Here $\mathbf U$ denotes a generic input to an axial update in the next-frame coordinate system, $\mathbf R^{\mathrm{recall}}$ the recalled evidence on the target grid, $\mathbf m$ its validity mask, and $\mathbf a_t$ the ego-motion condition from $t$ to $t+1$. Current and memory features use separate key and value projections, while their attention scores share one normalization so the target query can select between predicted content and observed evidence. For axis $a\in\{x,y\}$, $\mathbf Q_a$, $\mathbf L_a$, and $\mathbf A_a$ denote queries, attention scores, and normalized weights:
\begin{equation}
\begin{aligned}
&\mathbf Q_a=\operatorname{Query}_a(\mathbf U,\mathbf a_t),\\
&\mathbf L_a=\left[
\mathbf Q_aK_a^{\rm cur}(\mathbf U)^{\mathsf T},
\mathbf Q_aK_a^{\rm mem}(\mathbf R^{\mathrm{recall}})^{\mathsf T}
\right],\\
&\mathbf A_a=\operatorname{MaskedSoftmax}_a\!\left(
\frac{\mathbf L_a}{\sqrt{d_k}}+\mathbf B_a^{\rm pos};\mathbf m\right),\\
&\mathbf U\leftarrow\mathbf U+\operatorname{Proj}_a\!\left(\mathbf A_a
\begin{bmatrix}
V_a^{\rm cur}(\mathbf U)\\
V_a^{\rm mem}(\mathbf R^{\mathrm{recall}})
\end{bmatrix}\right),
\end{aligned}
\label{eq:supp_axial_interaction}
\end{equation}
where $\operatorname{Query}_a$ combines target content, position, and ego-motion. $V_a^{\rm cur}$ and $V_a^{\rm mem}$ are the value projections of current target features and recalled evidence, respectively. The key dimension per attention head is $d_k$, and $\mathbf B_a^{\rm pos}$ supplies relative position biases to both source blocks. $\operatorname{Proj}_a$ concatenates the attention heads and projects the aggregated features. The mask $\mathbf m$ excludes invalid memory positions.

Starting from the third cell's target state after historical updating, $\mathbf U_h=\mathbf U_{\rm loc}+\Delta\mathbf U_{\rm hist}$, the scene update applies the two axes sequentially ($x$ then $y$). A feed-forward residual update then produces $\mathbf U_g$, completing $\operatorname{SceneAttn}$ in the main paper. This factorization propagates recalled evidence over the full scene without constructing dense attention between all pairs of planar positions.

The remaining transition cells continue local spatial interaction and historical updates to produce $\widehat{\mathbf Z}_{t+1}$, corresponding to $\operatorname{Refine}$ in the main paper. During autoregressive inference, the model appends this prediction to the four-frame sliding history $\mathcal H_t$ and removes the oldest state:
\begin{equation}
\begin{aligned}
\mathcal H_{t+1}=\{&
(\bar{\mathbf Z}_{t-2},\mathbf P_{t-2}),
(\bar{\mathbf Z}_{t-1},\mathbf P_{t-1}),\\
&(\bar{\mathbf Z}_{t},\mathbf P_t),
(\widehat{\mathbf Z}_{t+1},\mathbf P_{t+1})\},
\end{aligned}
\label{eq:supp_history_update}
\end{equation}
where $\mathbf P_s\in\mathbb R^{4\times4}$ is frame $s$'s ego-to-global pose matrix. During inference, $\bar{\mathbf Z}_s$ is an encoded observation for $s\leq0$ or a previously predicted state for $s>0$. Memory $\mathcal M$ remains fixed throughout rollout. Training mixes encoded and predicted history states through scheduled sampling.

\begin{figure*}[t]
  \centering
  \includegraphics[width=0.96\textwidth]{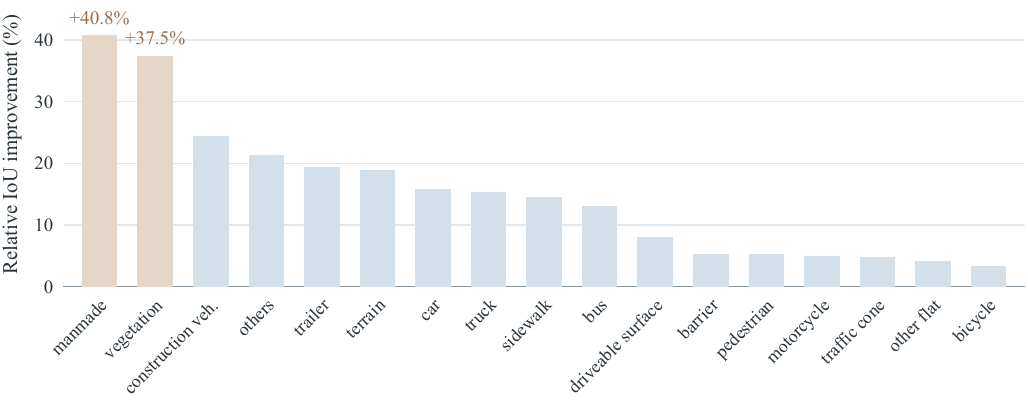}
  \caption{\textbf{Relative reconstruction-IoU gains of GeoTex over the class-wise mean of GenieDrive and I$^2$-World.}}
  \label{fig:supp_class_gain}
\end{figure*}

\section{Category-Wise Reconstruction Analysis}
\label{sec:supp_reconstruction_analysis}

The main paper analyzes the reconstruction ability of GeoTex Representation over structural interfaces and interiors. To further identify which semantic content benefits from GeoTex, we compute reconstruction IoU for each non-empty class on the complete validation set and use the class-wise mean of GenieDrive and I$^2$-World as the baseline. For each class, the relative gain is defined as
\begin{equation}
\begin{aligned}
\operatorname{Gain}={}&
\left(
\frac{\operatorname{IoU}_{\mathrm{GeoTex}}}
{\left(\operatorname{IoU}_{\mathrm{GenieDrive}}+
\operatorname{IoU}_{\mathrm{I^2\text{-}World}}\right)/2}
-1\right)\\
&\times100\%.
\end{aligned}
\label{eq:supp_class_gain}
\end{equation}

As shown in \cref{fig:supp_class_gain}, GeoTex improves every non-empty class, but the gains are not uniformly distributed. The largest improvements occur for \emph{manmade} and \emph{vegetation}, reaching 40.8\% and 37.5\%, respectively. These classes commonly contain continuous facades, irregular height variations, and structures extending across multiple spatial columns, making their reconstruction particularly dependent on within-column height patterns and cross-column relations. Classes with pronounced three-dimensional volume and contour variation also benefit substantially, including \emph{construction vehicle} at 24.4\% and \emph{trailer} at 19.4\%.

By contrast, the approximately planar \emph{driveable surface} and \emph{other flat} classes improve by only 8.0\% and 4.2\%. Gains for small-scale classes, including \emph{bicycle}, \emph{traffic cone}, \emph{motorcycle}, and \emph{pedestrian}, remain between 3.4\% and 5.4\%. The gains are mainly concentrated in categories with complex height variations and pronounced spatial extent across columns.

\section{Prediction-Domain Decoder Adaptation}
\label{sec:supp_decoder_adaptation}

During representation learning, the decoder receives latent representations encoded from observed occupancy, whereas forecasting requires it to decode autoregressively generated states. This distribution gap affects semantic classification and local boundary recovery. To examine its effect, we freeze each method's encoder and prediction module and adapt only the decoder using both encoded and predicted latent states. For GenieDrive, adaptation starts from our reproduced non-E2E forecasting model.

\begin{figure*}[!t]
  \centering
  \includegraphics[width=\textwidth]{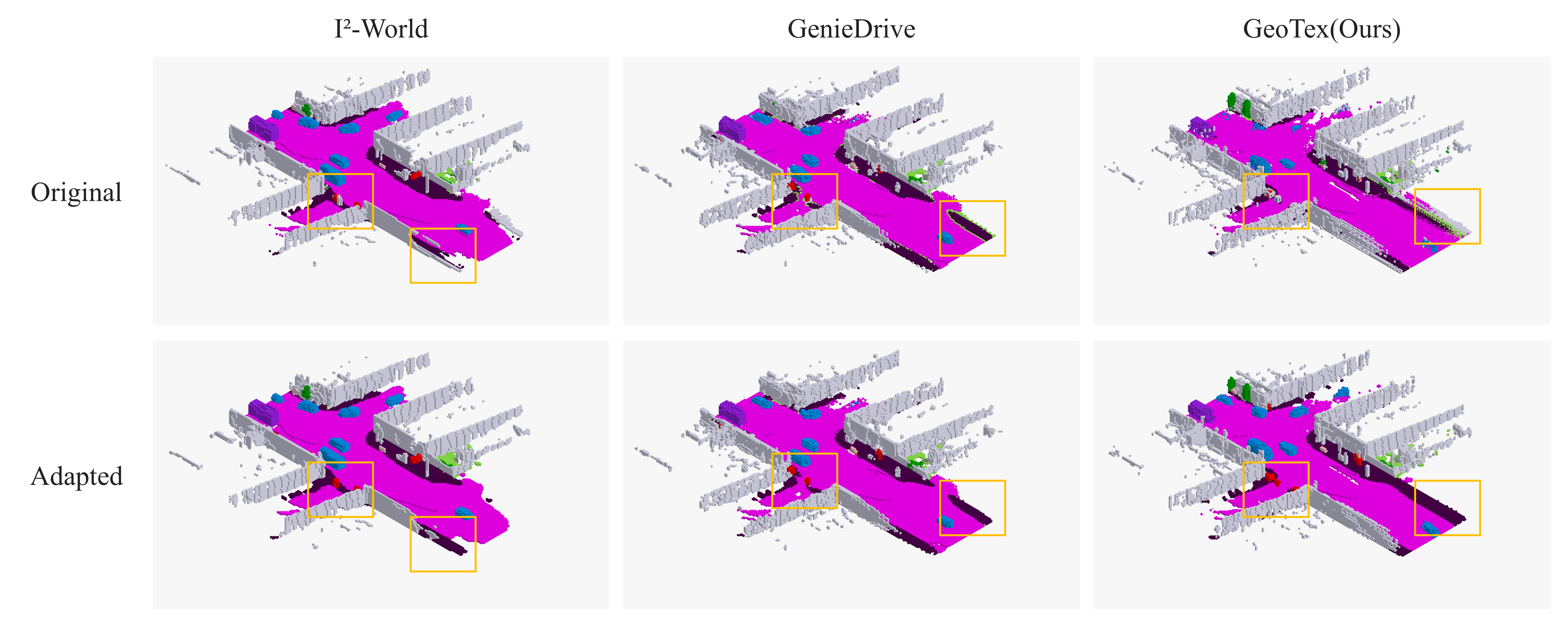}
  \caption{\textbf{Predictions before and after decoder adaptation across different methods.} The top and bottom rows use the original and adapted decoders, respectively. Yellow boxes highlight reductions in local semantic confusion, merging of neighboring classes, and grid-like occupancy artifacts.}
  \label{fig:supp_decoder_adaptation_comparison}
\end{figure*}

\begin{figure*}[!t]
  \centering
  \includegraphics[width=\textwidth]{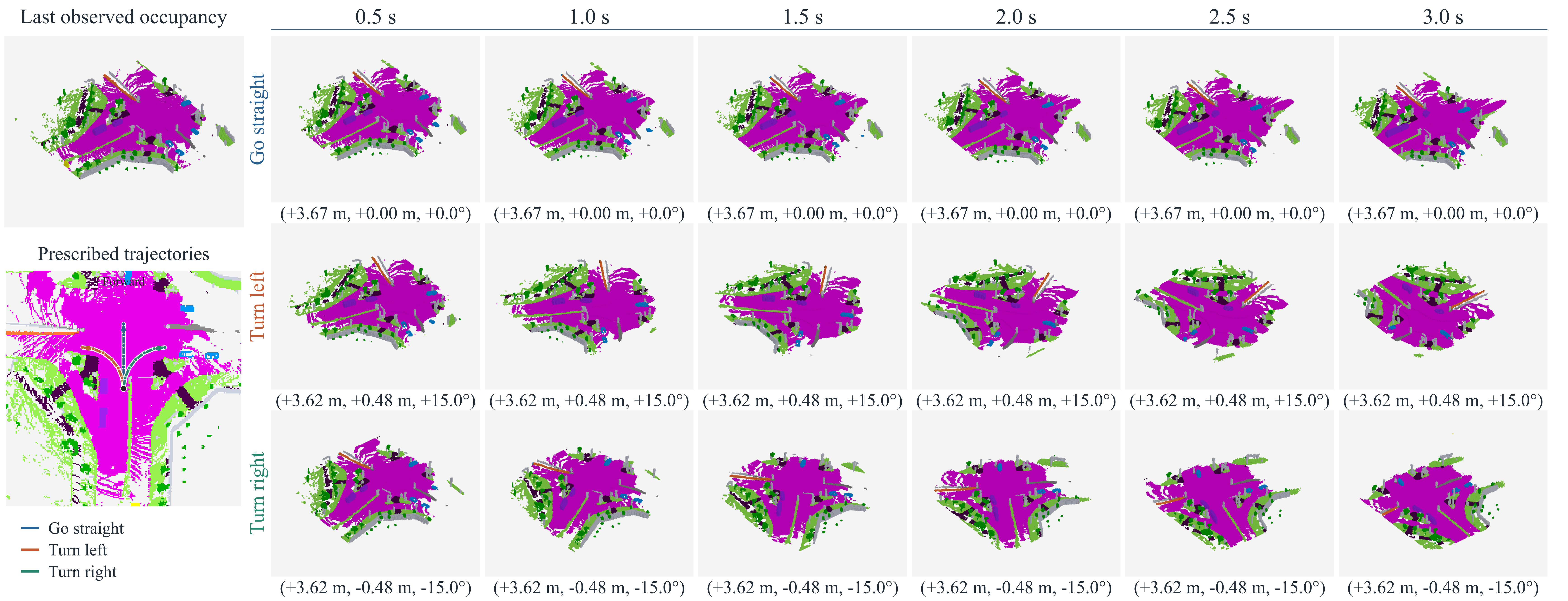}
  \caption{\textbf{Scene generation under ego-motion control.} The left column shows the last historical observation and the prescribed trajectories. The remaining columns show future occupancy sequences for going straight, turning left, and turning right. Below each prediction, $(\Delta x,\Delta y,\Delta\psi)$ denotes forward displacement, lateral displacement, and yaw change relative to the preceding ego frame.}
  \label{fig:supp_controllable_generation}
\end{figure*}

\begin{figure*}[!t]
  \centering
  \includegraphics[width=\textwidth,trim=0 135bp 0 0,clip]{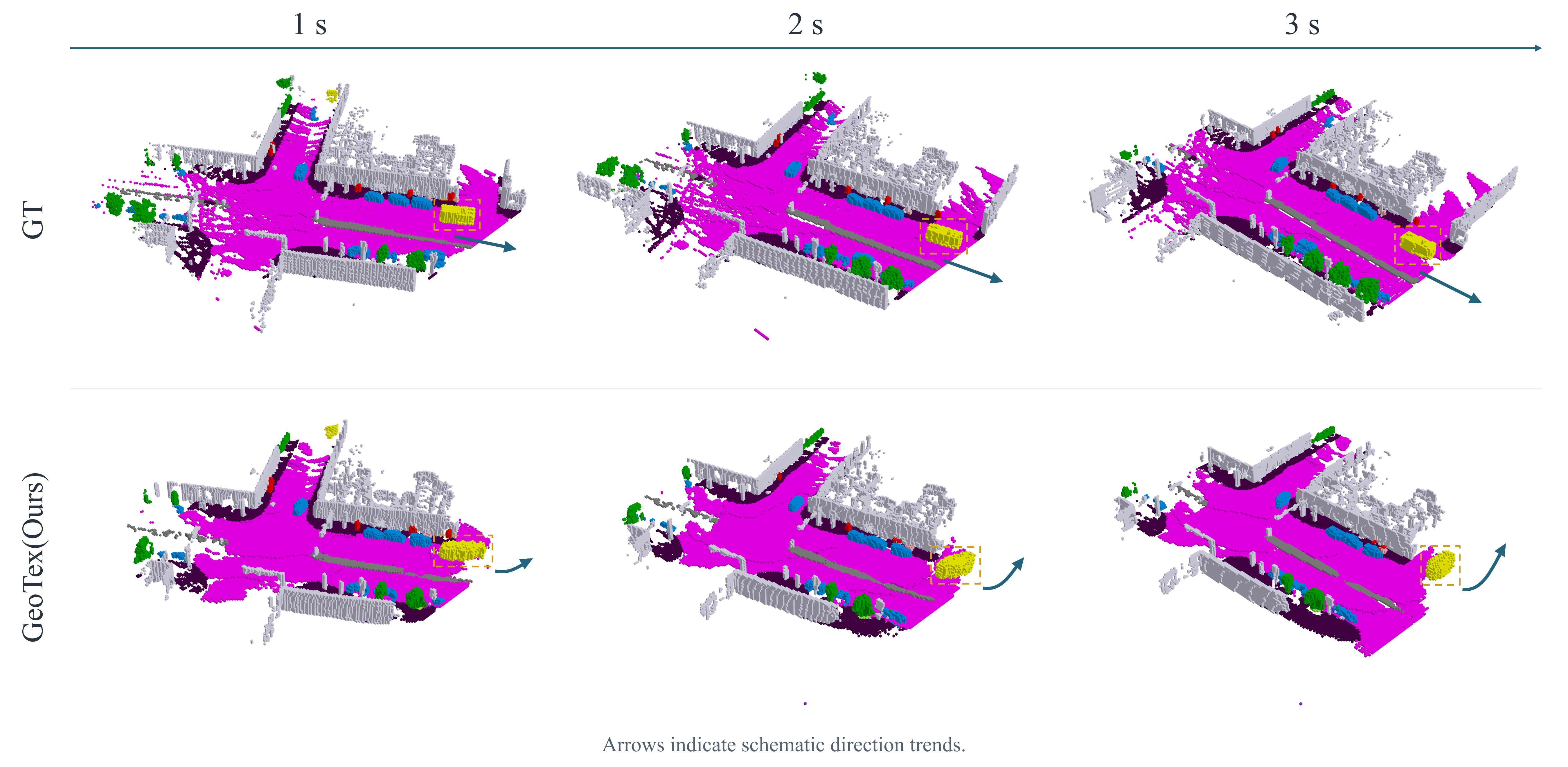}
  \caption{\textbf{Different future evolutions of a surrounding vehicle.} The yellow vehicle continues straight in GT, whereas GeoTex predicts a gradual turn toward a side road. This example highlights the distinction between matching the observed trajectory and assessing the plausibility of a future evolution.}
  \label{fig:supp_dynamic_future_ambiguity}
\end{figure*}

\begin{table}[t]
  \centering
  \caption{\textbf{Effect of prediction-domain decoder adaptation on Occ3D-nuScenes.} Avg. is the mean over the 1s, 2s, and 3s endpoints.}
  \label{tab:supp_decoder_adaptation}
  \footnotesize
  \setlength{\tabcolsep}{2.2pt}
  \renewcommand{\arraystretch}{1.08}
  \resizebox{\columnwidth}{!}{%
  \begin{tabular}{@{}ll|cccc|cccc@{}}
    \toprule
    \multirow{2}{*}{Method} & \multirow{2}{*}{Decoder} & \multicolumn{4}{c|}{mIoU (\%)$\uparrow$} & \multicolumn{4}{c}{IoU (\%)$\uparrow$} \\
    \cmidrule(lr){3-6}\cmidrule(lr){7-10}
    & & 1s & 2s & 3s & Avg. & 1s & 2s & 3s & Avg. \\
    \midrule
    \multirow{2}{*}{I$^2$-World~\cite{i2world}} & Original & 47.62 & 38.58 & 32.98 & 39.73 & 54.29 & 49.43 & 45.69 & 49.80 \\
    & Adapted & 49.27 & 40.57 & 35.11 & 41.65 & 54.98 & 50.32 & 46.70 & 50.67 \\
    \midrule
    \multirow{2}{*}{GenieDrive~\cite{geniedrive}} & Original & 48.29 & 38.61 & 32.79 & 39.90 & 55.46 & 49.52 & 45.18 & 50.05 \\
    & Adapted & 48.81 & 39.19 & 33.34 & 40.45 & 55.46 & 49.57 & 45.24 & 50.09 \\
    \midrule
    \multirow{2}{*}{GeoTex(Ours)} & Original & 47.72 & 38.87 & 32.85 & 39.81 & 60.72 & 55.89 & 51.61 & 56.07 \\
    & Adapted & \textbf{54.54} & \textbf{45.99} & \textbf{39.76} & \textbf{46.76} & \textbf{62.59} & \textbf{57.79} & \textbf{53.54} & \textbf{57.97} \\
    \bottomrule
  \end{tabular}
  }
\end{table}

As shown in \cref{tab:supp_decoder_adaptation}, all three methods improve in average mIoU. I$^2$-World increases from 39.73\% to 41.65\%, a relative gain of 4.83\%. Notably, the E2E experiment reported by GenieDrive~\cite{geniedrive} reduced I$^2$-World's average mIoU to 10.09\%. Unlike jointly updating the representation and forecasting model, decoder adaptation keeps the encoded and predicted latent states fixed and focuses optimization on decoding. I$^2$-World can therefore benefit from subsequent adaptation despite the degradation observed under that E2E setting.

GenieDrive gains 1.37\% in average mIoU while its IoU remains nearly unchanged. The yellow boxes in \cref{fig:supp_decoder_adaptation_comparison} illustrate these modest but clear local improvements: the green-class confusion along the road edge is reduced, and neighboring semantic regions become better separated. In the left highlighted region of I$^2$-World, the pedestrian previously merged with the building also becomes more distinct. These examples show that decoder adaptation reduces local semantic confusion and helps prevent small objects from being absorbed into neighboring structures. However, it only adjusts the decoding of fixed predicted states and does not alter the predicted latent-state sequence, so local corrections need not yield a large overall gain.

GeoTex shows the most pronounced decoding improvement. Its average mIoU increases from 39.81\% to 46.76\%, a relative gain of 17.46\%, while average IoU rises from 56.07\% to 57.97\%, a relative gain of 3.39\%. Before adaptation, its IoU already exceeds those of the other two methods, indicating that the model preserves scene occupancy structure well. The original decoder, however, does not fully translate this structural information into accurate voxel semantics.

In the illustrated scene, GeoTex's original decoder also produces conspicuous grid-like occupancy artifacts. The decoder first unfolds the compact planar state into height features and then progressively upsamples them into 3D occupancy. When applied to predicted states, a mapping learned from encoded observations can amplify inconsistencies between neighboring latent cells, producing regular voxel fragments and interleaved semantic labels. After adaptation, the grid-like confusion in the right highlighted region is substantially reduced, boundaries become more continuous, and the small object in the left region becomes more clearly separated from surrounding structures. Since the predicted latent states remain identical before and after adaptation, these improvements identify decoder mismatch as an important source of the artifacts and demonstrate that adaptation makes better use of the scene information already present in those states.

Overall, all three methods show larger relative gains in mIoU than in IoU, indicating a more pronounced benefit in semantic discrimination. The qualitative results also reveal corrections to occupancy boundaries and fine structures. For GeoTex, the high pre-adaptation IoU indicates that predicted states retain scene occupancy structure, while the subsequent mIoU gain shows that decoder adaptation improves their semantic readout.

\section{Ego-Motion-Controlled Scene Generation}
\label{sec:supp_controllable_generation}

We hold the four historical observations and travel speed fixed in an intersection scene and vary only the future turning direction. Straight, left-turn, and right-turn ego-motion sequences each drive a six-frame rollout at 2\,Hz, covering the next 3 seconds. The same trained model generates all three sequences without further training.

As shown in \cref{fig:supp_controllable_generation}, the main road extends along its original direction under straight-driving control. Under left- and right-turn controls, the intersection and surrounding structures change continuously in the ego-centric frame as the vehicle turns, producing distinct future occupancy sequences. The main road contours and roadside structures remain coherent while the model responds to the control inputs. GeoTex Prediction uses ego-motion to locate historical content relative to future target positions. Observation memory supplies scene context, and recent history provides complementary state updates. Thus, the same historical scene supports different motion-conditioned futures while retaining coherent structure.

\section{Dynamic Future Ambiguity and Evaluation Limitations}
\label{sec:supp_dynamic_future_ambiguity}

\Cref{fig:supp_dynamic_future_ambiguity} shows a divergence between GeoTex and GT in the evolution of a surrounding vehicle. The yellow vehicle continues along the road in GT, whereas GeoTex predicts a gradual turn toward a side road. GeoTex Prediction updates scene content using historical evidence after geometric alignment, so object evolution is not restricted to rigid propagation. In this scene, the road branch provides spatial scope for an alternative maneuver. Limited historical observations may not uniquely determine a vehicle's future intent, so predictions that differ from the observed trajectory also need to be assessed against scene and motion constraints.

This example illustrates a limitation of evaluation against a single observed future. Both mIoU and IoU measure voxel overlap with GT. For relatively stable structures such as roads and buildings, they directly reflect geometric and semantic agreement. Dynamic objects, however, may admit several plausible future behaviors. Even a trajectory that follows the road layout and maintains motion continuity is penalized when its position or orientation differs from GT. These metrics therefore measure fidelity to the observed future, but cannot independently distinguish implausible motion from a plausible alternative.

Our future work will explore evaluation protocols that account for both prediction accuracy and generative plausibility. Alongside voxel-level geometry and semantics, these protocols should consider motion continuity, road constraints, interaction consistency, and coverage of multiple possible futures, providing a more complete assessment of scene evolution in occupancy world models.

\end{document}